\documentclass[conference]{IEEEtran}
\IEEEoverridecommandlockouts
\usepackage{cite}
\usepackage{amsmath,amssymb,amsfonts}
\usepackage{algorithmic}
\usepackage{graphicx}
\usepackage{textcomp}
\usepackage{xcolor}
\def\BibTeX{{\rm B\kern-.05em{\sc i\kern-.025em b}\kern-.08em
    T\kern-.1667em\lower.7ex\hbox{E}\kern-.125emX}}
\begin{document}

\title{A Text Block Refinement Framework For Text Classification and Object Recognition From Academic Articles \\
}

\author{Li Jinghong, Ota Koichi, Gu Wen, Hasegawa Shinobu\\
Japan Advanced Institute of Science and Technology, Japan\\
\{s2220006,ota,wgu,hasegawa\}@jaist.ac.jp\\
}

\maketitle

\begin{abstract}
With the widespread use of the internet, it has become increasingly crucial to extract specific information from vast amounts of academic articles efficiently. Data mining techniques are generally employed to solve this issue. However, data mining for academic articles is challenging since it requires automatically extracting specific patterns in complex and unstructured layout documents. Current data mining methods for academic articles employ rule-based(RB) or machine learning(ML) approaches. However, using rule-based methods incurs a high coding cost for complex typesetting articles. On the other hand, simply using machine learning methods requires annotation work for complex content types within the paper, which can be costly. Furthermore, only using machine learning can lead to cases where patterns easily recognized by rule-based methods are mistakenly extracted. To overcome these issues, from the perspective of analyzing the standard layout and typesetting used in the specified publication, we emphasize implementing specific methods for specific characteristics in academic articles. We have developed a novel Text Block Refinement Framework (TBRF), a machine learning and rule-based scheme hybrid. We used the well-known \textit{ACL proceeding} articles as experimental data for the validation experiment. The experiment shows that our approach achieved over 95\% classification accuracy and 90\% detection accuracy for tables and figures.
\end{abstract}

\begin{IEEEkeywords}
Text mining, Machine learning, Text Block Refinement, Object detection.
\end{IEEEkeywords}

\section{Introduction}

With the advancement of information science, an overwhelming number of scientific articles are available. Despite their importance to researchers and students as a critical source of advanced knowledge and reference material, selecting and understanding important information from this vast quantity is becoming increasingly difficult \cite{1}. Hence, Text analysis tools for academic articles have been increasingly developed, including natural language processing techniques like extractive automatic summarization, abstract automatic summarization, and visualization slide generation\cite{2}.

In order to carry out the task mentioned above, it is important to efficiently collect and organize the language text elements (main text, itemized form, sections, footnotes) and non-language text elements (figures, tables, formulas, quotation marks) within academic articles \cite{jsai}. For this issue, data mining techniques are fundamental approaches to classify and identify information in the article into different categories \cite{3}. Document layout analysis (DLA) purposes to detect and annotate the physical structure of documents \cite{5}. However, parsing the layout and analyzing the content of academic articles can be challenging and intricate. The layout of research articles is often irregular, and the typesetting styles vary depending on the publication \cite{6}. Therefore, there is a tendency to extract discontinuous data when extracting internal information. For instance, The flow of the main narrative from a file may be broken in mid-sentence by errors derived from the reading order of individual text blocks and interruptions, including figure captions, footnotes, and headers \cite{7}. Rule-based conditional branching and regular expressions are conventional methods to process text to solve the above issue, such as parsing sentences and identifying keywords \cite{7}. However, these methods could be improved in their ability to process the complex structure used in academic articles, which may contain many patterns and irregularities that are not easily detected by traditional algorithms. To address this issue, researchers have turned to machine learning algorithms, which are more flexible and able to learn and adapt based on the presented data. Machine learning methods can take into account the specific patterns in academic articles and improve the accuracy and effectiveness of automatic text processing \cite{8}. However, academic articles may have varying fonts and typesetting, which require more complex vectorization and annotation methods to increase generality. Formulating these intricate techniques can be time-consuming and expensive.

\begin{table*}[htbp]
\centering
\caption{The summary of previous work}
\begin{tabular}{llll}
\hline
\textbf{Reference}                                                             & \textbf{Method}                                                  & \textbf{Feature/Advantage}                                                                                                                                                 & \textbf{Limitations}                                                                                                                                                              \\ \hline
\textit{\begin{tabular}[c]{@{}l@{}}Table Tranformer (2022)\cite{table tran}\end{tabular}} & \begin{tabular}[c]{@{}l@{}}Detection \\ Transformer\end{tabular} & \begin{tabular}[c]{@{}l@{}}Large training data , and supports multiple \\ input modalities and is useful for many \\ modeling approaches\end{tabular}                      & \begin{tabular}[c]{@{}l@{}}Miss detection in complex pattern \\ matching involving continuous tables.\end{tabular}                                                                \\ \hline
\textit{\begin{tabular}[c]{@{}l@{}}Table net (2020)\cite{table net}\end{tabular}}           & \begin{tabular}[c]{@{}l@{}}Deep\\ learning\end{tabular}          & \begin{tabular}[c]{@{}l@{}}This model uses interdependence between \\ table detection and structure recognition tasks \\ to segment table and column regions.\end{tabular} & Header of the table is difficult to fit                                                                                                                                           \\ \hline
\textit{\begin{tabular}[c]{@{}l@{}}Pdffigure2.0 (2016)\cite{pdffigure}\end{tabular}}        & Rule-based                                                       & \begin{tabular}[c]{@{}l@{}}Analyzes page structure and locates figures \\ and tables by analyzing empty regions within text.\end{tabular}                                  & \begin{tabular}[c]{@{}l@{}}Miss detection occured when figures \\ and tables appear continuously\end{tabular}                                                                     \\ \hline
\textit{\begin{tabular}[c]{@{}l@{}}Tabula (2018)\cite{tabula}\end{tabular}}              & Rule-based                                                       & \begin{tabular}[c]{@{}l@{}}It uses customizable heuristics to detect tables\\ and reconstruct cell structure based on text \\ and ruling lines in the PDF.\end{tabular}    & \begin{tabular}[c]{@{}l@{}}If recognition is not restricted to the table zone, \\ the body-text and section title patterns may \\ be mixed with the detected result.\end{tabular} \\ \hline
\textit{\begin{tabular}[c]{@{}l@{}}Grobid (2008--2023)\cite{grobid}\end{tabular}}        & CRF                                                              & \begin{tabular}[c]{@{}l@{}}Extract and reorganise not only the content \\ but also the layout and text styling information.\end{tabular}                                   & \begin{tabular}[c]{@{}l@{}}it is difficult to extract noise-free text \\ because information with figures, tables,\\  and equations are included in the body-text.\end{tabular}   \\ \hline
\end{tabular}
\end{table*}

This study aims to develop a TBRF system to automate the processing of multi-modal elements in specific academic articles using a small-scale training module. Specifically, by utilizing the layout features of line spacing and column spacing in academic articles, we divide the document into text blocks, refine their raw data characteristics, and construct a framework for recognizing specific objects within the articles by classifying and merging text blocks. Our approach combines rule-based and machine learning methods to detect object features. The contributes of TBRF system mainly in the following four aspects.\\
\textbf{1.} We constructed an integrated encoding template that reflects information about text blocks in the articles, including size, coordinates, font type, and font size.\\
\textbf{2.} We present a method for constructing a small scale dataset for element classification in specific academic publications. The feature of our method is that it involves human annotation of both linguistic and non-linguistic text information in the articles, and can be done in a short amount of time.\\
\textbf{3.} We used machine learning techniques to automatically classify the multi-modal elements within the article, using the constructed dataset from \textbf{2.} to verified the learning performance by small-scale training data.\\
\textbf{4.} Based on the classification results, we performed tuning to element block to construct an object recognition module for detecting the element's type. To verify the effectiveness of our proposed method for object recognition, we conducted comparison experiments with existing multi-modal document processing methods.\\

\section{Related work}
The summary of related work on data mining for academic articles is shown in \textbf{Table I}. Previous studies have been limited by their unstable distinguishability of text blocks that appear within figures and tables. In this study, we aim to overcome the limitations by refining text blocks based on various features, such as position, size, line and column spacing, font type, and font size. These features enable us to better distinguish objects within the acadmic articles.

\section{methodology}

This section introduces the Text Block Refinement Framework for Object Recognition from Scientific articles (TBRF). The framework focuses on the contents presented in \textbf{Table II} for object recognition. TBRF aims to recognize object characteristics by combining machine learning and rule-based methods. We aim to develop new bottom-up level templates that reflect the features of text blocks using less training data. This will improve the efficiency of human annotation work for machine learning and reduce the time required for coding work in the rule-based scheme. We use a 3-phase approach: Acquisition of raw data, Rule-based detection for apparent characteristics, and Object Recognition for unobvious characteristics by machine learning work. The overview of our model is shown in \textbf{Fig.1}. In the following subsections, we provide a comprehensive description of each component of the proposed approach.

\begin{figure}[htbp]
\centering
\includegraphics[width=8cm,height=5cm]{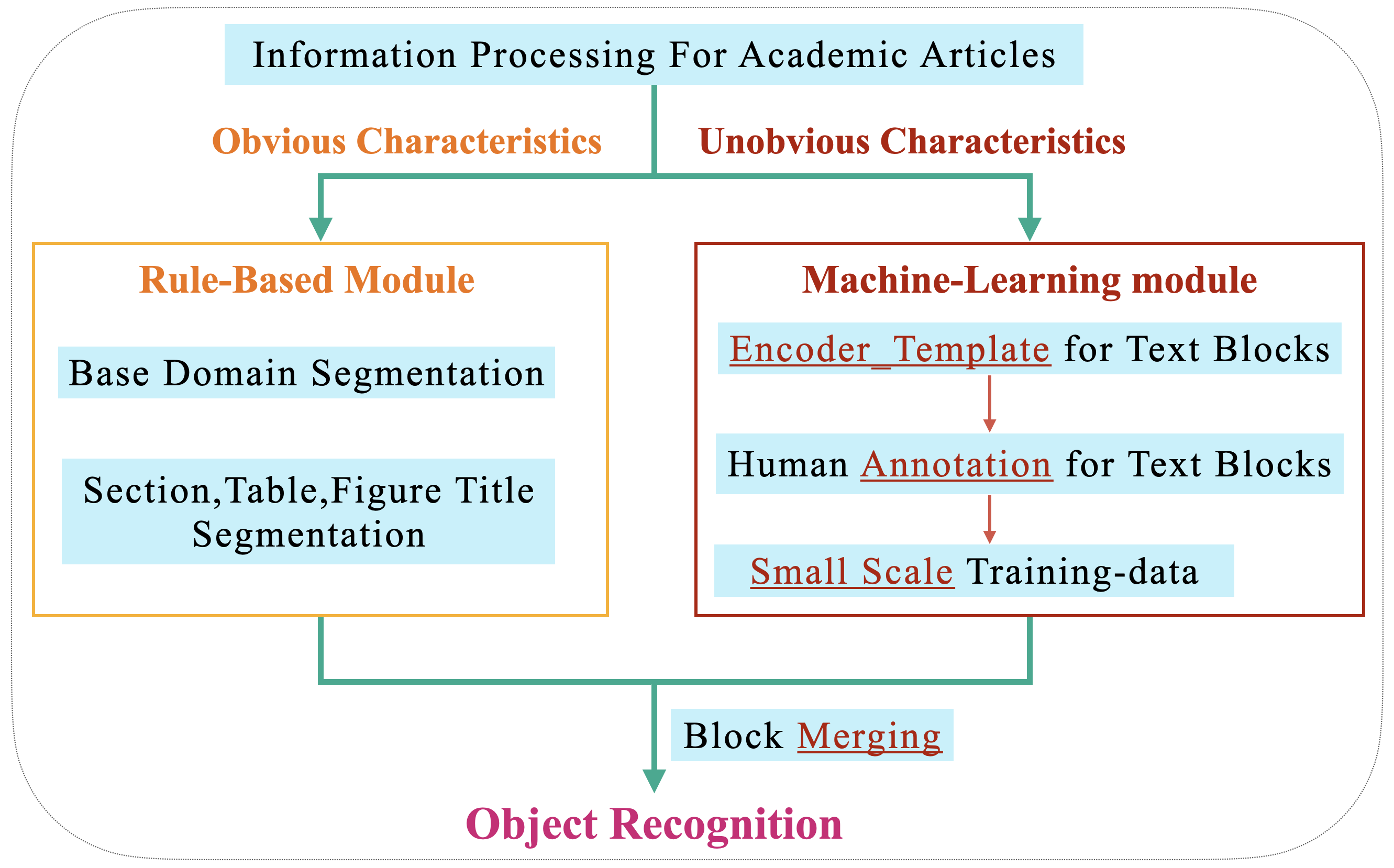}

\caption{Overview}
\end{figure}


\subsection{Definition of Object Types}

Academic articles contain various types of objects, such as figures, tables, equations, etc. Objects are able to be classified into three classes based on their usage: Body-text (text-based sentences), Supplementary information (visual data to help readers understand the article), and Accessory information (additional information related to the article). The corresponding element for each type is shown in \textbf{Table II}. \textbf{Table III} shows the elements targeted in this research and how they were detected.

\begin{table}[htbp]
\centering
\caption{Definition of Object Types}
\begin{tabular}{|c|c|}
\hline
\textbf{Object Type}                                                       & \textbf{Element}                                                                                                                     \\ \hline
\textbf{Body-text}                                                            & Sentence group in body of article                                                                                                    \\ \hline
\textbf{\begin{tabular}[c]{@{}c@{}}Supplementary\\  information\end{tabular}} & \begin{tabular}[c]{@{}c@{}}Figure and Table zones, \\ Figure and Table titles  \\ Equations, Algorithms, Sections title\end{tabular} \\ \hline
\textbf{Accessory information}                                                & \begin{tabular}[c]{@{}c@{}}Page number, Footnote, \\ Meta information\end{tabular}                                                   \\ \hline
\end{tabular}
\end{table}

\begin{table}[htbp]
\caption{Elements and their detection method}
\begin{tabular}{|l|l|l|l|}
\hline
\textbf{Element}                                                           & \textbf{Type} & \textbf{Reason}                                                                                                                          & \textbf{Method} \\ \hline
\textit{\textbf{\begin{tabular}[c]{@{}l@{}}Section \\ Title\end{tabular}}} & Obvious       & \begin{tabular}[c]{@{}l@{}}- Continuity of section numbers \\ - Specific fonts for section titles\end{tabular}                             & RB              \\ \hline
\textit{\textbf{\begin{tabular}[c]{@{}l@{}}Tab\&Fig\\ Title\end{tabular}}} & Obvious       & \begin{tabular}[c]{@{}l@{}}The format of Tab/Fig titles is \\ generally consistent within the \\ same academic publication.\end{tabular} & RB              \\ \hline
\textit{\textbf{Body-Text}}                                                & Unobvious     & Discontinuous data                                                                                                                       & ML              \\ \hline
\textit{\textbf{Figure}}                                                   & Unobvious     & Irregular text block included                                                                                                            & ML              \\ \hline
\textit{\textbf{Table}}                                                    & Unobvious     & Irregular text block included                                                                                                            & ML              \\ \hline
\textit{\textbf{Page\_Num}}                                                & Unobvious     & Similar text block in tables                                                                                                             & ML              \\ \hline
\textit{\textbf{Footnote}}                                                 & Unobvious     & Similar text block in body-text                                                                                                          & ML              \\ \hline
\end{tabular}
\end{table}

\vspace{-3mm}
\subsection{Text Block Raw-data acquisition}

Text in academic papers appears in various places, such as within the bodytext, figures, tables, and footnotes. It can be organized into text blocks based on line and column spacing to determine the types of objects where this text appears. Secondary processing can be performed based on the characteristics of these text blocks. This section aims to extract the objects' characteristics in \textbf{Table II} by obtaining features such as font, size, and location on the page of these text blocks.

\subsubsection{\textbf{Text block Parsing}}
Scientific repositories currently store research articles in PDF format. The first step is to focus on parsing the text blocks within PDF articles. This can be automated using the external library \textit{pymupdf}\cite{pymupdf} in Python. By analyzing the raw structure of academic articles, the line and column spacing, characteristics can be used to divide the text of a PDF file into multiple sub-text blocks. These blocks contain specific content, and their combination results compose the unstructured page layout, as depicted in \textbf{Fig.2}. The "Page.get\_text(“blocks”)" method of "pymupdf" can be used to extract the text blocks of each page.

\begin{figure}[htbp]
\centering
\includegraphics[width=7.5cm,height=4.7cm]{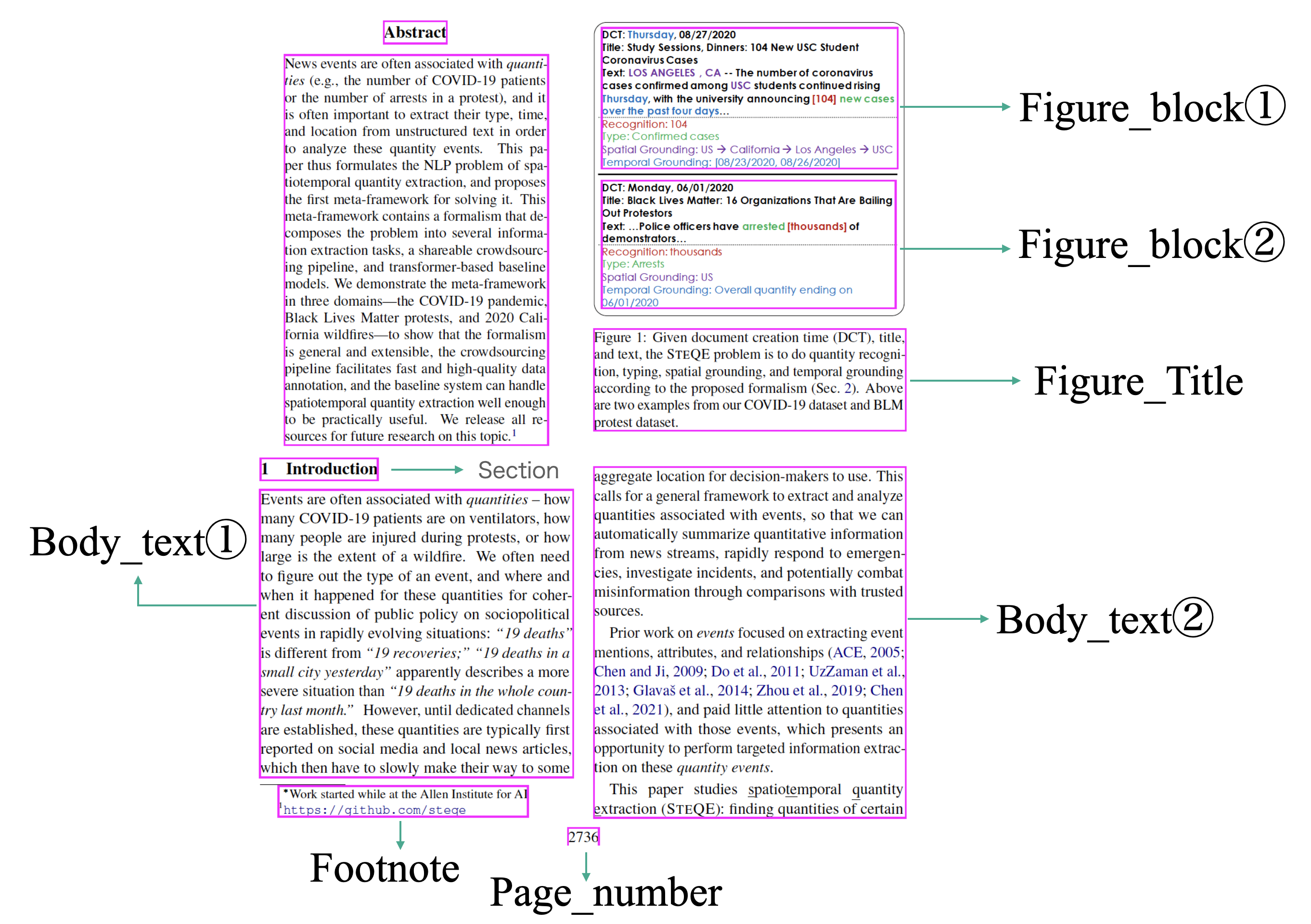}
\caption{unstructured page layout: Sample article\cite{block}}
\end{figure}

\subsubsection{\textbf{Extracting accompanying information from text}}

When extracting text blocks using Pymupdf, the accompanying text's information, such as font, size, and style, can also be obtained \cite{pymupdf}. This information can be used in various ways, such as analyzing the document's structure, identifying headers and footers, and detecting text requiring special formatting. Analyzing these characteristics makes it possible to optimize the extraction process and improve the accuracy of extracting specific features. For example, font size can be used to identify the main-sections of the document or to differentiate between headings and body-text. Font style can be used to detect emphasis or to identify quotes. Moreover, utilizing the accompanying text's information enables data extraction in a more organized and structured manner, resulting in more easily obtainable, low-noise output.

\subsection{Rule-based detection for obvious characteristics}

The internal content of academic papers is classified into four categories: Basic Information Domain, Body Domain, Reference Domain, and Appendix Domain, with rule-based detection specific to the Base Domain and section titles. This study focuses on processing academic articles with specific section numbers assigned. We explore each area by matching the beginning string with regular expressions, except for the appendix area. This segmentation ensures that each Base Domain is adequately organized and presented clearly and concisely. Patterns of Section titles are divided into two types:\textbf{\textit{main-section}} and \textbf{\textit{sub-section}}. Each match method uses regular expressions to fit section numbers.

\subsection{Object recognition for unobvious characteristics }

According to the procedure of \textbf{Fig.1}, creating a machine learning dataset for object recognition involves performing vector encoding and annotation work. This step considers features such as text, font, text block size and coordinates in academic articles. Next, we use a highly versatile SVM(Support Vector Machine) classifier to obtain the corresponding class for each text block. Based on the results of the text block, we develop a block merging algorithm that considers the continuity of the matching text blocks and recognizes the zone of each object within the area it is contained.


\subsubsection{\textbf{Encoder template}}

Eight vector elements are selected to create a vector that reflects the characteristics of an article. Each vector element is constructed to embed the characteristics of a specific object. The encoding method for each element is as follows:\\
\textbf{1-2, Left\_Position and Right\_Position: }
As the body-text block is often Justify Align, the goal is to accurately calculate the location of the beginning of each block and obtain common characteristics of the text. Expressly, we set the Left/Right-aligned block as a left/right boundary line and calculate the division between each block's left/Right coordinate and the boundary coordinate.\\
\begin{equation}
\small
    Code_{(left|right)} = \frac{Block\_coordinate_{(Left|Right)}}{Boundary\_coordinate_{(Left|Right)}} 
\end{equation}
\textbf{3-4, Top\_Position and Bottom\_Position: }
Footnote and page number information is often displayed at the bottom of the page, so it is necessary to encode that information to distinguish it from other information. The encoding method is the same as the one described in Left /Right Position but switches from left/right to upper/lower.\\
\begin{equation}
\small
    Code_{(top|bottom)} = \frac{Block\_coordinate_{(top|bottom)}}{Boundary\_coordinate_{(top|bottom)}} 
\end{equation}
\textbf{5-6, Width\_Length and Height\_Length: }
While the blocks of body-text are distributed more regularly, it is necessary to grasp the width and height characteristics of the blocks to recognize the supplement information blocks, as they are irregularly distributed and have many small blocks. Therefore, in an article, set the largest width/height as the standard for all blocks, perform a division of the width/height of each block, and encode it.\\
\begin{equation}
\small
    Code_{(width|height)} = \frac{Block\_size_{(width|height)}}{Max\_size_{(width|height)}} 
\end{equation}
\textbf{7, Font\_Type(\textbf{ft}): }
In most cases, the body-text font has the highest frequency in academic articles. Therefore, following the font acquisition method in \textbf{B.}., all blocks are scanned to aggregate the corresponding character count for each font, and the font type with the highest frequency is considered the body-text font. Next, the most frequently appearing font in each block is calculated, and if it is the body-text font, it is encoded as' 1'. Otherwise, it is encoded as' 0'.\\
\begin{equation}
\small
    Body_{font} = Index\_of[Max\{\sum{ft1},...,\sum{ftN}\}] 
\end{equation}
\begin{equation}
\small
Code_{ft} = 
    \begin{cases}
        {\textbf{1} \ (Body\_font)}\\
        {\textbf{0} \ (Others)}
    \end{cases}
\end{equation}
\textbf{8, Font\_Size(\textbf{fs}): }
First, set the standard font size to the most frequently appeared font size among the body fonts. Next, determine the most frequent font size in each block and set it as the font size for that block. Finally, divide each block's font size by the standard font size and encode it. In equation (7), fss means font size in that text block.\\
\begin{equation}
\small
    Body_{fs} = Index\_of [Max\{\sum{fs1},...,\sum{fsN}]
\end{equation}
\begin{equation}
\small
    Block_{fs} = Index\_of [Max\{\sum{fss1},...,\sum{fssN}]
\end{equation}
\begin{equation}
\small
    Code_{fs} =  \frac{Block_{fs}}{Body_{fs}} 
\end{equation}
\subsubsection{\textbf{Annotation for Type of contents}}
Humans annotate text blocks by judging the information of body-text, supplementary information, and accessories related to the text block based on the description of \textbf{Table I}. Combined with vectors constructed in \textbf{1)}, a dataset for object recognition is constructed.
\subsubsection{\textbf{Classifier}}
Support Vector Machine (SVM) is one of the classifiers that boast particularly high accuracy in machine learning. SVM has the advantage of high applicability even to unknown data \cite{SVM}. It seeks to maximize the margin when drawing a boundary line to divide the training data. Maximizing the margin improves the classification accuracy for unknown data, and the generalization performance becomes higher. Therefore, SVM has high adaptability and versatility for academic articles with complex structure combinations. For instance, some special cases of text blocks in figures have the same font and size as the body-text. The SVM classifier's margin and tolerance range setting are considered to improve training effectiveness to recognize these cases.

\subsubsection{\textbf{Determine the zone of figures and charts by block merging}}
Based on the classification results obtained through training, each block is recognized as either body-text information, supplementary information, or accessory information label. In this section, we combine the specified consecutive blocks based on the supplementary information block, adjust them with the position of the figure and table titles, and determine the zone of the figures and tables. 


Although the writing style of figure and table titles may depend on the publication, we specify the \textit{ACL proceeding} format in this article and perform pattern matching. Firstly, match the figure/table title strings with the regular expression. Next, identify the text block indicating the position of the matched figure/table title. Then, combine the consecutive supplementary information blocks just before (Prior) and just behind the figure/table title by getting the frame of the text block groups. Finally, referring to the pdffigure2.0 method \cite{pdffigure}, If the width of the figure/table area is larger, it is determined as the figure/table area. If the block's width where the labeled title is located is larger, resize the figure/table area to match that width. The results of the match samples are shown in \textbf{Fig.3}.

\begin{figure}[htbp]
\centering

  \begin{minipage}[b]{0.45\linewidth}
    \centering
    \includegraphics[width=4cm,height=4.5cm]{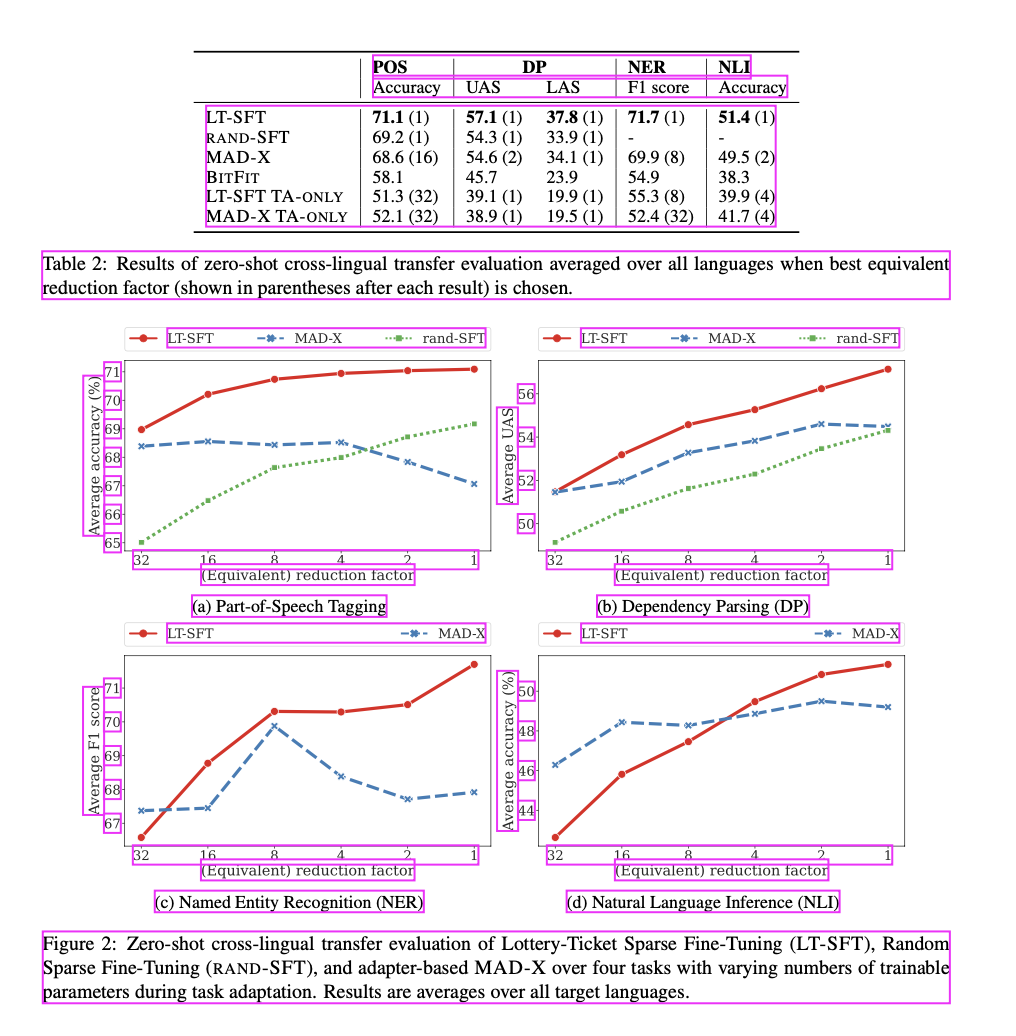}
    \caption{(a)unaligned layout and text block}
  \end{minipage}
  \begin{minipage}[b]{0.45\linewidth}
    \centering
    
    \includegraphics[width=4cm,height=4.5cm]{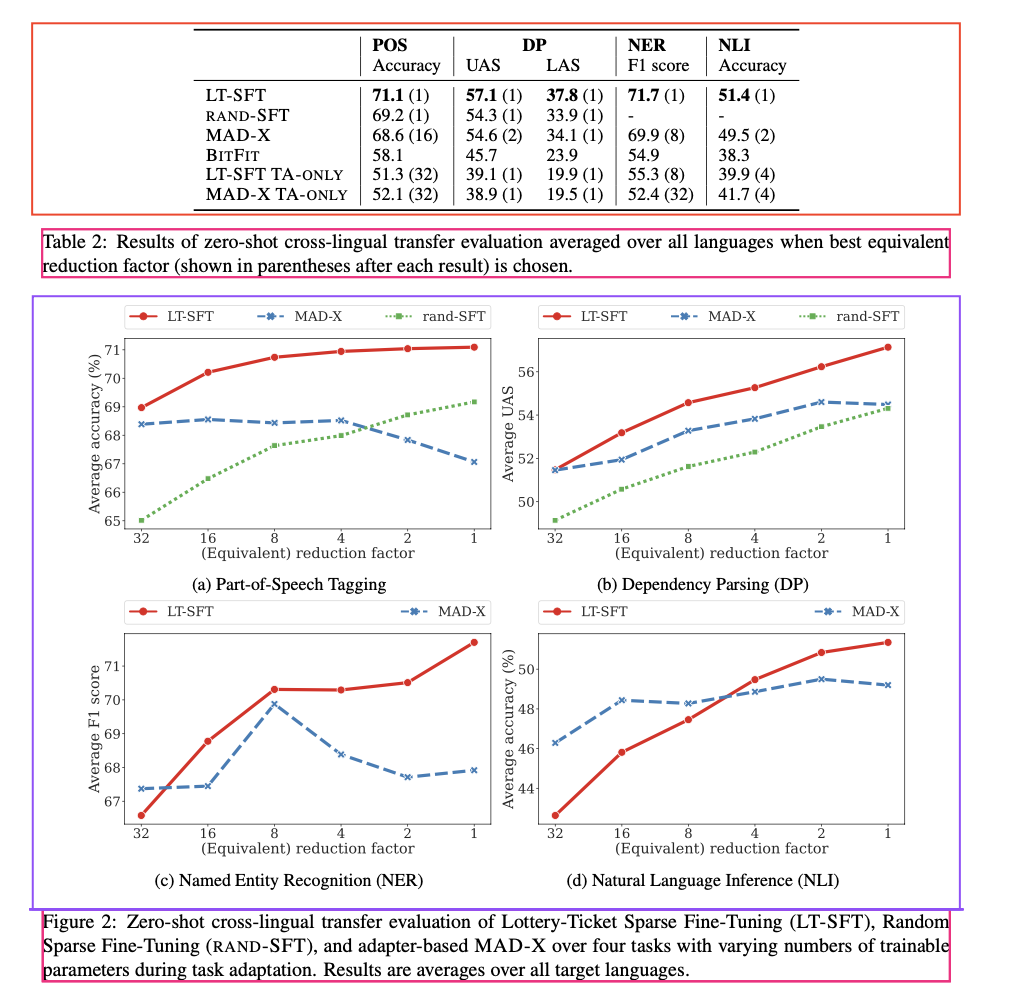}
    
    \caption{(b)Figure.Table zone detection}
  \end{minipage}
  \caption{Sample of Object Detection: Sample article\cite{detect}}
\end{figure}

\section{Experiment: Text block Classification}

\subsection{Article collection of PDF format}

We collected 768 articles in PDF format from \textit{Proceedings of the 60th Annual Meeting of the Association for Computational Linguistics}. to use as our experimental data.

\subsection{Experirment}
To test the learning of our small data, we randomly selected ten articles from 768 research articles to build a training module. We encoded and annotated them with human input and constructed a dataset that took approximately an hour to complete. For the data split, 90\% were set as training data and 10\% as validation data. The label distribution is shown in \textbf{Table IV}.

\begin{table}[htbp]
\centering
\caption{Details of training\&validation dataset}
\begin{tabular}{|c|c|c|}
\hline
\textbf{Class}&\textbf{Number of labels}&\textbf{Ratio}\\
\hline
Body-text&514&33.9\%\\
\hline
Supplement&894&58.9\%\\
\hline
Accessory&109&7.2\%\\
\hline
Total&1518&-\\
\hline
\end{tabular}
\end{table}


Through a process of trial and error, we have determined the optimal training parameters for the Support Vector Machine (SVM) model, which are presented in \textbf{Table V}. These parameters were selected based on a rigorous evaluation of various combinations of hyperparameters, including the kernel type, regularization parameter, and degree of the polynomial kernel. Our results demonstrate that these parameters significantly improve the performance of the SVM model, resulting in more accurate and reliable predictions.


\begin{table}[htbp]
\centering
\caption{Details of SVM Parameter}
\begin{tabular}{|c|c|c|l|}
\hline
\textbf{Parameter}  & \textbf{Value} & \textbf{Description}                                                                                         & \textbf{Degree}        \\ \hline
C     & 100            & \begin{tabular}[c]{@{}c@{}}There are fewer misclassification \\ points in the decision area.\end{tabular} & Large                  \\ \hline
Gamma & 0.1            & \begin{tabular}[c]{@{}c@{}}The decision boundary is\\  a simple decision boundary.\end{tabular}              & Small                  \\ \hline
Rbf   & -              & \begin{tabular}[c]{@{}c@{}}It can represent a\\  non-linear boundary\end{tabular}                            & \multicolumn{1}{c|}{-} \\ \hline
\end{tabular}
\end{table}

\subsection{Evaluation}
The training and validation dataset underwent random processing in 100 experiments to verify result stability. Each experiment was designed to reduce the impact of random fluctuations and provide a more accurate representation of the experiment result. By examining the dataset, we can ensure reliable results. From the 100 times' experimental results, we derived the average value and statistics for each label's accuracy, precision, recall, and F1-score obtained through Support Vector Machines (SVM), as presented in \textbf{Table VI}. We achieved an overall precision, recall, and F1-score of over \textbf{95\%} in classifying types of text blocks. These results provide a fine basis for the block merging in object recognition experiment.

\begin{table}[htbp]
\centering
\caption{SVM Result of validation data}
\begin{tabular}{|c|c|c|c|c|}
\hline
\textbf{Result}&\textbf{All label}&\textbf{Body-text}&\textbf{Supplement} &\textbf{Accessory}\\
\hline
Precision&0.974&0.935&0.987&0.938\\
\hline
Recall&0.954&0.980&0.952&0.991\\
\hline
F1-Score&0.963&0.957&0.964&0.963\\
\hline
\end{tabular}

\end{table}
\section{Experiment: Object recognition}
Using 20 randomly selected articles in PDF format as test data for object recognition. Initially, the sampling data in section \textbf{IV} is used as input for the SVM classifier to build a training module. After obtaining the classification result, a comparison experiment is conducted to verify the effectiveness of the Table and Figure detection. The comparison is made with previous studies such as Pdffigure2.0, Tabula, Table Transformer, and Table net.

\subsection{Evaluation}
 The classification results of 20 articles via SVM are shown in \textbf{Table VII}. As our result, it was verified that the text block classification for specific academic articles can be handled with our small amount of training data.
\begin{table}[htbp]
\centering
\caption{SVM Result of test data}
\label{t7}
\begin{tabular}{|c|c|c|c|c|}
\hline
\textbf{Result}&\textbf{All label}&\textbf{Body-text}&\textbf{Supplement} &\textbf{Accessory}\\
\hline
Precision&0.973&0.959&0.990&0.906\\
\hline
Recall&0.951&0.981&0.974&0.970\\
\hline
F1-Score&0.963&0.970&0.982&0.937\\
\hline
\end{tabular}
\end{table}

\begin{table*}[htbp]
\centering
\caption{Object recognition result(Number of Figure: 68, Number of Table: 93)}
\begin{tabular}{|l|l|c|c|c|c|}
\hline
\textbf{}                        & \textbf{Training data scale}                                                & \multicolumn{1}{l|}{\textbf{Accepted Figure}} & \multicolumn{1}{l|}{\textbf{Accuracy Figure}} & \multicolumn{1}{l|}{\textbf{Accepted Table}} & \multicolumn{1}{l|}{\textbf{Accuracy Table}} \\ \hline
\textit{\textbf{Our approach}}   & 10 articles                                                                 & \textbf{64/68}                                & \textbf{0.940}                                & 85/93                                        & 0.913                                        \\ \hline
\textit{Table Tranformer (2022)} & \begin{tabular}[c]{@{}l@{}}948K Tables in \\ 100,000s articles\end{tabular} & -                                             & -                                             & 85/93                                        & 0.913                                        \\ \hline
\textit{Table net (2020)}        & \begin{tabular}[c]{@{}l@{}}509 Images for \\ academic articles\end{tabular} & -                                             & -                                             & 53/93                                        & 0.569                                        \\ \hline
\textit{Pdffigure2.0 (2016)}     & Rule-based                                                                  & 61/68                                         & 0.897                                         & \textbf{90/93}                               & \textbf{0.967}                               \\ \hline
\textit{Tabula (2018)}           & Rule-based                                                                  & -                                             & -                                             & 22/93                                        & 0.236                                        \\ \hline
\end{tabular}
\end{table*}
After finishing block merging processing, the object recognition accuracy was obtained. The comparison experiment results are shown in \textbf{Table VIII}. We achieved competitive recognition results for figures and tables in specific academic publications with only a small amount of training data from 10 papers. Eventually, we conduct a result analysis with high-performance models - Table Transformer (TATR) and PDFfigure 2.0.\\
\textbf{- Compare with Table Transformer: }According to the experimental results, we have achieved the same recognition accuracy as the advanced module "Table Transformer" in terms of table area recognition. As shown in \textbf{Fig.4}, our method demonstrated higher stability than Table Transformer in complex pattern matching involving continuous tables.\\
\textbf{- Compare with Pdffigure2.0: }The experiment showed that Pdffigure2.0 had trouble distinguishing between figure and body-text when multiple consecutive text boxes were in the image. We improved the accuracy of figure recognition by using font type and size to differentiate text boxes from body-text, as shown in \textbf{Fig.5}. However, our approach did not reach PDFfigure2.0's recognition accuracy for table because we did not consider the whitespace between text boxes when recognizing tables. To improve it, we need to encode whitespace between text blocks and effectively detect table characteristics.\\   

\begin{figure}[htbp]
\centering
  \begin{minipage}[b]{0.45\linewidth}
    \centering
    \includegraphics[width=4cm,height=4.2cm]{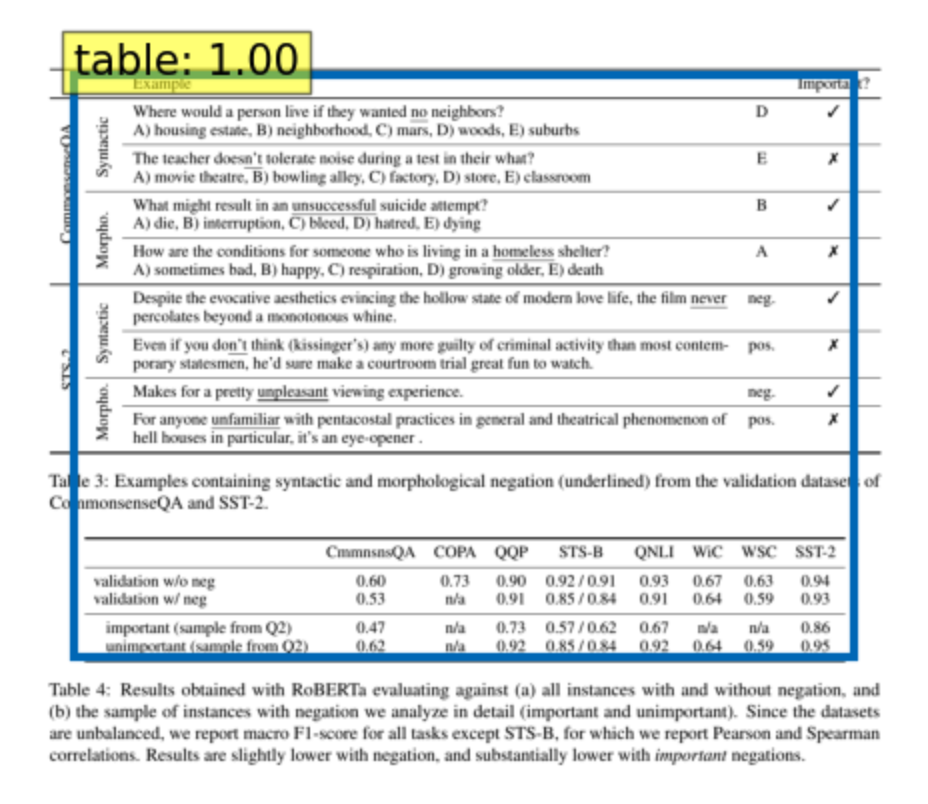}
    \caption{(a)Table Transformer}
  \end{minipage}
  \begin{minipage}[b]{0.45\linewidth}
    \centering
    
    \includegraphics[width=4cm,height=4.2cm]{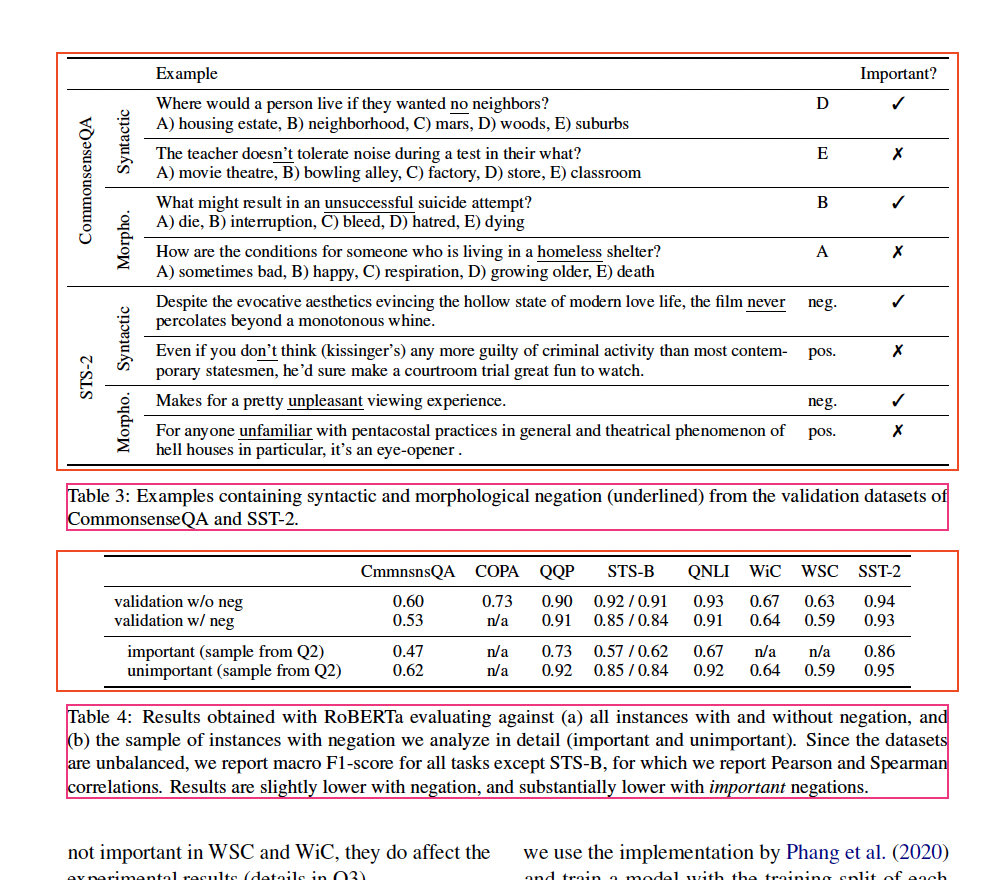}
    
    \caption{(b)Our approach}
  \end{minipage}
  
  \caption{Sample of Comparing with Table Transformer: Sample article\cite{compare1}}
\end{figure}

\begin{figure}[htbp]
\centering

  \begin{minipage}[b]{0.45\linewidth}
    \centering
    \includegraphics[width=4cm,height=3cm]{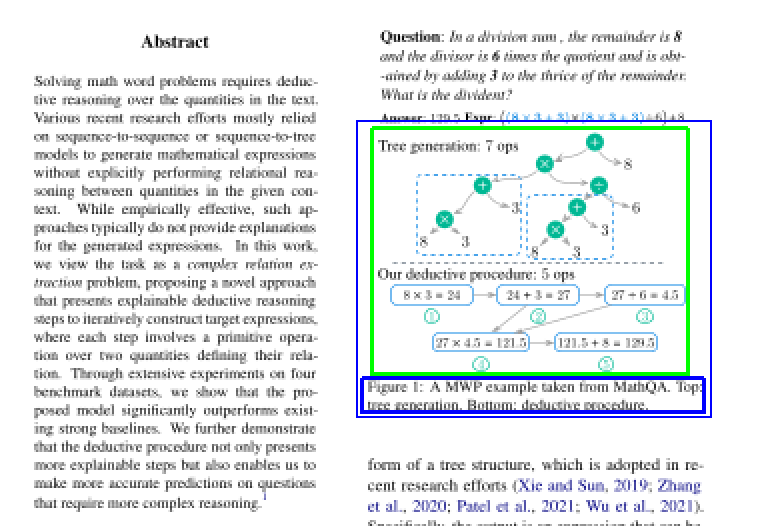}
    \caption{(a)Pdffigure2.0}
  \end{minipage}
  \begin{minipage}[b]{0.45\linewidth}
    \centering
    
    \includegraphics[width=4cm,height=3cm]{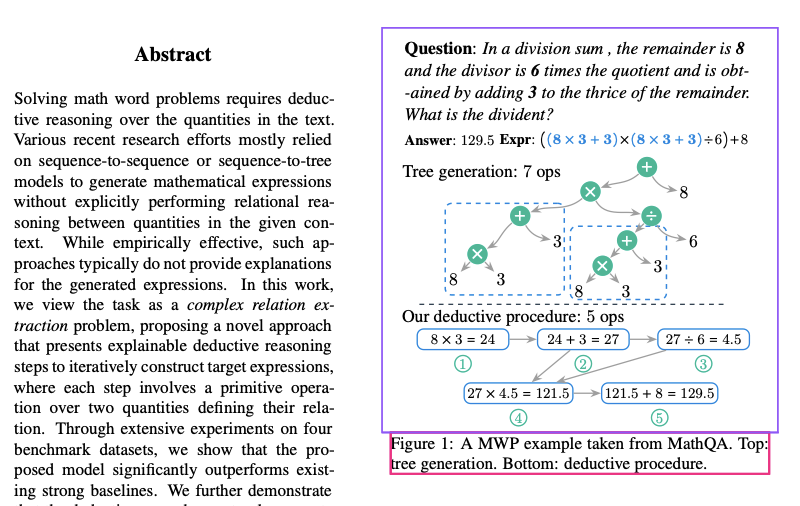}
    
    \caption{(b)Our approach}
  \end{minipage}
  
  \caption{Sample of Comparing with Pdffigure2.0: Sample article\cite{compare2}}
\end{figure}
\section{Conclusion}
This study aims to recognize objects in academic PDF articles. The framework developed for this purpose is called Text Block Refinement Framework for Object Recognition from Scientific article (TBRF). This framework uses rule-based and regular expression recognition to identify domains and section information. It also employs a bottom-up level encoded template for machine learning, which utilizes accompanying information of text blocks to identify complex body-text, accessories, figures, and tables. The effectiveness of this framework was demonstrated through the use of a small scale training dataset and an SVM classifier. A specialized text block fusion algorithm was also developed to extract figures and tables using the classification results of SVM, achieving an accuracy of over 90\%. Overall, experimental results showed the effectiveness of this framework.\\
\textbf{Future tasks to improve this framework include the following:}\\
\textbf{(1) Adding types of object: }One of the future challenges is to recognize patterns of equations,itemized form, algorithm areas, and lists within the body-text.\\
\textbf{(2) Improvement of Encode Template :} In this study, we created eight features based on text font, size, block size, and position as additional information about text blocks. By including these features, our templates can better capture document complexity and provide more accurate recognition results. However, obtaining text styles like italic and bold, and the distance between text blocks is challenging and necessary for more sophisticated templates. These improvements to the Encode Template will help develop more reliable and accurate document recognition systems.\\
\textbf{(3) Constructing datasets for natural language processing task :} Few studies have examined the link between figures, tables, and body-text in academic articles, which play a crucial role in natural language processing tasks like navigators and automatic summaries. By analyzing the text in figure and table areas using object recognition and matching it to the text's main content, we can automatically generate datasets that contain a network of knowledge information, linking related information. This could significantly benefit natural language processing and have important implications for education and computer visualization.

\end{document}